\documentclass[runningheads]{llncs}
\usepackage[T1]{fontenc}
\usepackage{graphicx}
\usepackage{amsmath} 
\usepackage{amsfonts}
\usepackage{booktabs}
\usepackage{placeins}
\usepackage{bm}
\usepackage[misc]{ifsym} 
\usepackage[colorlinks=true]{hyperref}

\begin{document}
\title{Multi-modal Data Binding for Survival Analysis Modeling with Incomplete Data and Annotations}
\titlerunning{Survival Analysis Modeling with Incomplete Data and Annotations}
%
\author{
Linhao Qu$^*$\inst{1} \and
Dan Huang$^*$\inst{2} \and
Shaoting Zhang\inst{1} \and
Xiaosong Wang$^{(\textrm{\Letter})}$\inst{1}}

\institute{Shanghai Artificial Intelligence Laboratory, Shanghai, China \and 
Department of Pathology, Fudan University Shanghai Cancer Center, Shanghai, P.R. China. 270 Dong An Road, Shanghai 200032, China. \\ \email{wangxiaosong@pjlab.org.cn}}

\authorrunning{L. Qu \textit{et al.}}
%
%
\renewcommand{\thefootnote}{}
\footnotetext{$^*$Linhao Qu and Dan Huang contributed equally.}

\maketitle              
\begin{abstract}
Survival analysis stands as a pivotal process in cancer treatment research, crucial for predicting patient survival rates accurately. Recent advancements in data collection techniques have paved the way for enhancing survival predictions by integrating information from multiple modalities. However, real-world scenarios often present challenges with incomplete data, particularly when dealing with censored survival labels. Prior works have addressed missing modalities but have overlooked incomplete labels, which can introduce bias and limit model efficacy. To bridge this gap, we introduce a novel framework that simultaneously handles incomplete data across modalities and censored survival labels. Our approach employs advanced foundation models to encode individual modalities and align them into a universal representation space for seamless fusion. By generating pseudo labels and incorporating uncertainty, we significantly enhance predictive accuracy. The proposed method demonstrates outstanding prediction accuracy in two survival analysis tasks on both employed datasets. This innovative approach overcomes limitations associated with disparate modalities and improves the feasibility of comprehensive survival analysis using multiple large foundation models. 

\keywords{Multi-modality  \and Survival Analysis \and Missing Data.}
\end{abstract}
\section{Introduction}

\begin{figure}[t]
  \centering
   \includegraphics[width=\linewidth]{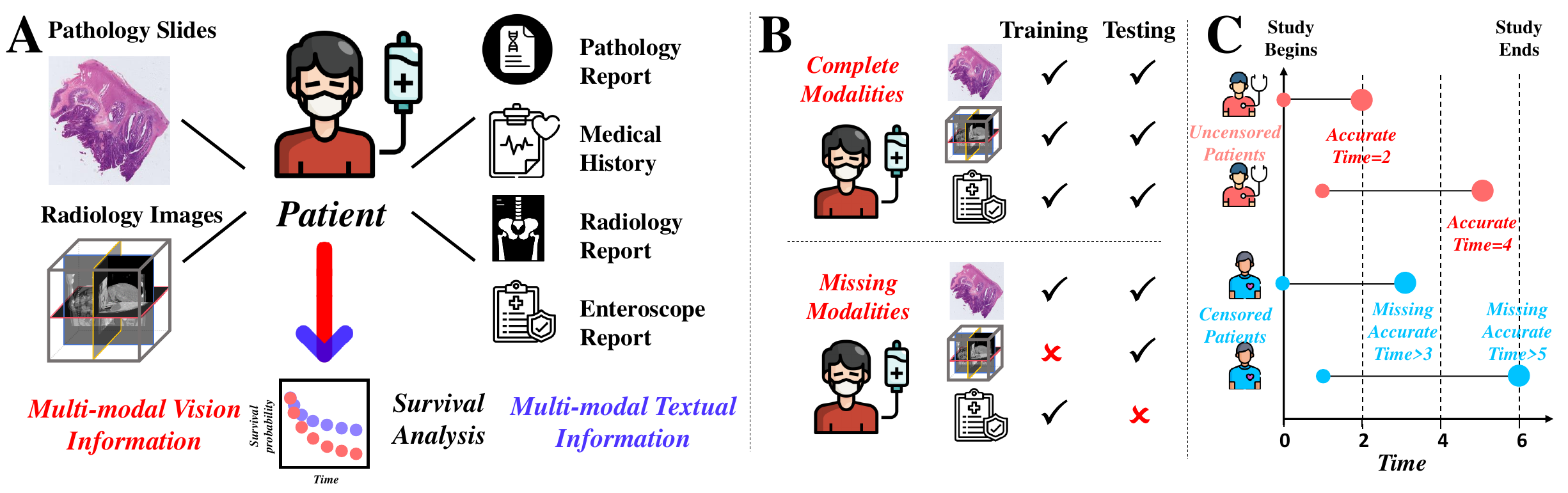}
   \caption{Overview of the problem definition. (A) Employing multi-modal vision and textual data for patient-wise survival analysis. (B) Missing modality issues during training and testing. (C) Missing accurate label issues for censored patients.}
   \label{fig:intro}
\end{figure}

Survival analysis is a vital process in cancer treatment research, enabling the prediction of important outcomes such as patient survival rates \cite{zhu2017wsisa,chen2020pathomic,chen2021pan,courtiol2019deep,yao2019deep}. Recent advancements~\cite{shen2016surviv,kashyap2019quantitative} in data collection techniques have opened new avenues for improving the accuracy of survival predictions by leveraging information from multiple modalities. An accurate prediction of the survival rate and time could primarily facilitate the precise composition of treatment planning. 

The fusion of information from multiple modalities has been the recent trend of research in survival analysis. By leveraging cutting-edge techniques (e.g., cross-attention~\cite{zhou2023cross} and co-attention~\cite{chen2021multimodal}) from the field of vision and language, it is popular to jointly train a model with two data modalities, e.g., pathology images and genomics. Nonetheless, these methods often require high data integrity and are constrained by limitations involving more than two modalities, as shown in Fig.~\ref{fig:intro} A and B.
Moreover, in real-world survival analysis, it is typical to encounter a significant proportion of right-censored data, where the event of interest has not occurred or remains unknown by the end of the follow-up period, as is defined as missing accurate label issues shown in Fig.~\ref{fig:intro} C.

To address this challenge, previous works first tackle the incompleteness of input data in multi-modal survival analysis. These methods \cite{cui2022survival,hou2023hybrid} focus on synthesizing the missing modalities, leveraging statistical techniques and generative model-based reconstruction algorithms to estimate the missing values. By filling in the missing information, these approaches strive to ensure that the analysis is conducted on as complete information as possible. Nonetheless, they still overlook the equally important issue of incomplete labels, which may introduce bias and limit the training efficacy of survival models. There is an urgent need for innovative approaches that account for both incomplete modalities and labels.

In this work, we demonstrate a joint framework that revolutionizes survival analysis by handling incomplete data across modalities and censored survival labels together. First, we take advantage of the current availability of advanced foundation models capable of encoding individual modalities. Then, the computed multi-modal embeddings are bound into a universal representation space via multi-modal feature alignment, paving the way for a seamless fusion of diverse modalities in a missing modality setting. Unlike previous methods, our method goes beyond merely handling missing modalities, instead addressing the formidable challenge of incomplete survival labels. By generating pseudo labels and incorporating uncertainty in the training of censored data, we significantly improve the predictive accuracy of survival prediction. This innovative approach eliminates the limitations associated with disparate modalities and enhances the feasibility of conducting comprehensive survival analysis. Crucially, our method also allows for clear interpretability of each modality's importance via an explicit attention mechanism.

The contribution of the proposed multi-modal survival analysis framework is three-fold: 1) We proposed a multi-modal survival analysis framework by considering the incompleteness in both the input data and label; 2)  we utilize a variety of foundation models to encode each modality and bind them into aligned representations for a more generalizable means of multi-modal data fusion; 3) We demonstrate outstanding prediction accuracy in two survival tasks across two real clinical datasets.

\begin{figure}[t]
  \centering
   \includegraphics[width=\linewidth]{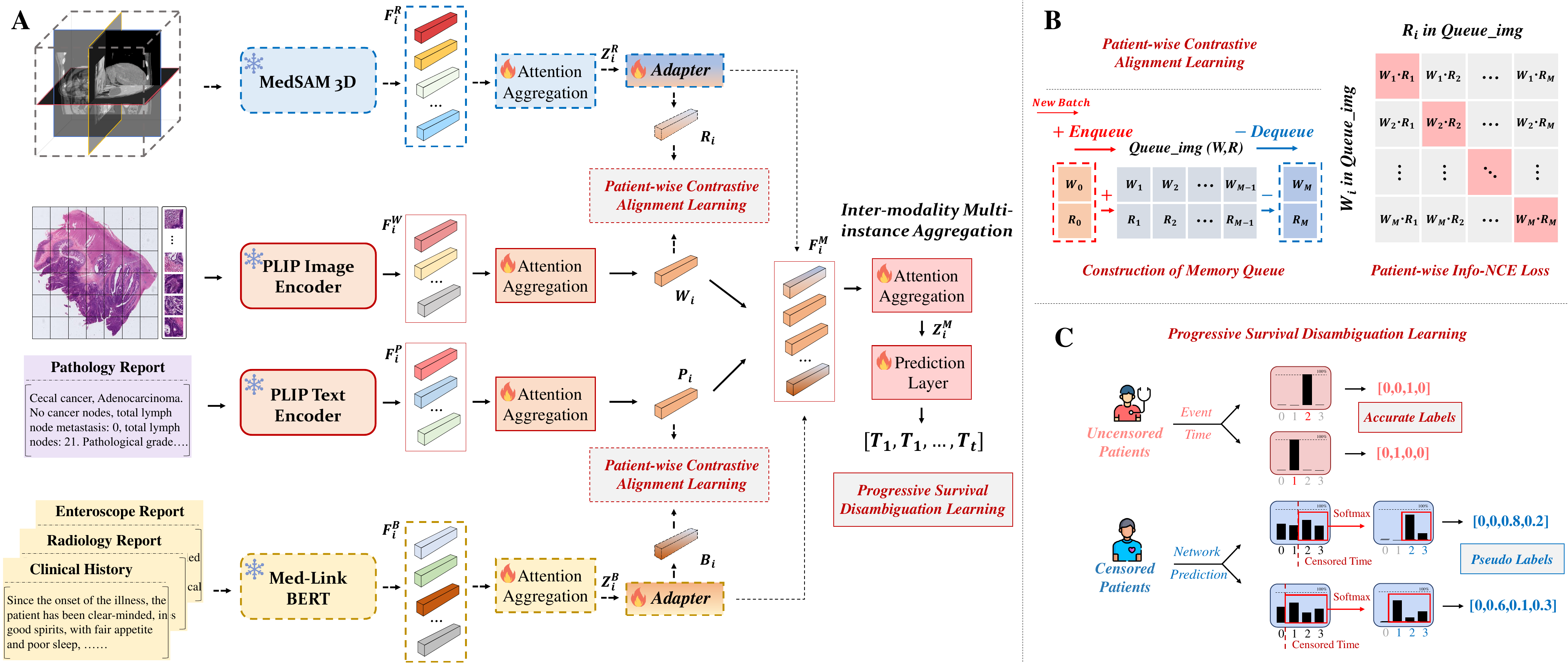}
   \caption{(A) Overview of the proposed framework. Solid lines: constant modalities; dashed lines: potentially missing modalities. (B) Diagram of Patient-wise Contrastive Alignment Learning. (C) Diagram of Progressive Survival Disambiguation Learning.}
   \label{fig:overview}
\end{figure}

\section{Method}

The overall training framework is shown in Fig.~\ref{fig:overview}. A. We first categorize all modalities into four types: radiology images, pathology Whole Slide Images (WSIs), pathology reports, and other clinical notes. Then, we utilize pre-trained modality-specific Foundation Models (FMs) to extract features from each modality separately (see Sec.~\ref{MSFM}); next, we introduce attention-based multi-instance feature aggregation for both intra-modality and inter-modality (see Sec.~\ref{MIAggreg}). To learn a single joint embedding space that encompasses patient-specific modalities, we propose a Patient-wise Contrastive Alignment Learning based on the Adapter and Contrastive Learning techniques (see Sec.~\ref{PWCA}). 
Finally, we introduce Progressive Survival Disambiguation Learning to address the issue of estimating unknown interval risks for censored patients (see Sec.~\ref{PSDL}). The overall loss function of the framework is defined as $
L=\lambda L_{\text {con }}+L_{\text {surv }}
$, where $L_{con}$ denotes the contrastive loss (see Sec.~\ref{PWCA}), $L_{surv}$ denotes the survival loss (see Sec.~\ref{PSDL}), and $\lambda$ denotes the balancing weight coefficient.

\subsection{Encoding Modalities with Modality-specific Foundation Models}
\label{MSFM}
\textbf{Multi-modal Input Data}: This paper includes a variety of important modalities for survival analysis of colorectal cancer patients, which are: (1) pathology WSIs; (2) radiology images such as CT or MRI; (3) pathology reports: descriptions of tumors regarding the patient's pathology slides, including tumor location, shape and size, grading, infiltration, invasion, and the status of major genetic targets; (4) radiology reports: descriptions of tumors corresponding to the patient's radiology images; (5) clinical history; (6) colonoscopy reports. All patients have at least one WSI and a corresponding pathology report, while the presence of other modalities can vary with certain incompleteness.

A single general FM may not be adequate for comprehensive encoding all modalities due to the diversity in format and content. For instances, FMs trained predominantly on natural image datasets fall short in accurately interpreting radiologic anatomical structures. Models trained on pathological images may not grasp the nuances of radiological anatomy. To overcome these limitations, we leverage modality-specific FMs, enhancing encoding precision. 

\noindent\textbf{Pathology Images and Reports}:
PLIP~\cite{huang2023visual} is applied to extract visual features from WSIs and textual features from pathology reports. Each WSI, after background removal, is divided into $n_W$ patches of 224$\times$224 at 10x magnification. PLIP's Image Encoder outputs a 1$\times$512 visual feature vector for each patch and together forms a set sized $n_W\times$512. The patch count per WSI can vary, and for individuals with multiple WSIs, the vectors are concatenated to form an extensive WSI feature set ${F}_{{i}}^{{W}} \in \mathbb{R}^{N_i^W \times 512}$, where $N_i^W$ is the total patch count from all WSIs of patient $i$. Pathology reports are segmented into sections by keywords (avoiding length limit of FM) and each section is encoded into a 1$\times$512 vector by PLIP's Text Encoder, creating a feature set ${F}_{{i}}^{{P}} \in \mathbb{R}^{N_i^P \times 512}$, with $N_i^P$ representing the section count, which varies by patient. These keywords used to segment the reports were provided by pathologists through structured text report data, including tumor location, size, grading, etc.

\noindent\textbf{Radiology Images}: For 3D radiology images (CT or MRI), the pre-trained MedSAM-3D \cite{lei2023medlsam} is utilized. Initially, each image is segmented into $n_R$ 3D patches of size 128$\times$128$\times$128 via sliding window. Subsequently, MedSAM-3D maps these patches to 1$\times$512 feature vectors, as an array with dimensions of $n_R$$\times$512. For individuals with variable number of images, feature vectors are concatenated into an array ${F}_{{i}}^{{R}} \in \mathbb{R}^{N_i^R \times 512}$, where $N_i^R$ denotes the aggregate number of 3D patches from all radiographic images for patient $i$.

\noindent\textbf{Other Clinical Notes}: 
The pre-trained BioLinkBERT-large \cite{yasunaga2022linkbert} is employed to tokenize and encode other clinical notes, including radiology reports, medical history, and colonoscopy reports. 
These textual data are segmented by keywords and encoded into feature vectors. 
Embeddings from multiple reports are concatenated to form a comprehensive feature set ${F}_i^{{B}} \in \mathbb{R}^{N_i^B \times 1024}$, with $N_i^B$ indicating the total segment count across all clinical notes for a patient.

\subsection{Attention-based Intra-modality and Inter-modality Multi-instance Aggregation}
\label{MIAggreg}
In multi-modal analysis, the challenges include (1) the potential absence of multiple modalities during both training and testing, (2) the variable number of feature vectors within each modality, and (3) the necessity to quantify the importance of intra-modal and inter-modal factors on outcomes. Innovatively, we address these issues by unifying the aggregation of intra-modal and inter-modal features into a problem of multi-instance aggregation based on the attention mechanism, offering a flexible and efficient solution. 

\noindent\textbf{Intra-modality Aggregation}: 
We aggregate the patient-wise feature vector sets extracted from radiology, pathology, and other medical notes into a uniform-dimensional feature vector, respectively. Specifically, we take the feature vector set of radiology data ${F}_{i, j}^{{R}}$ as an example, where the instances being aggregated are the feature vectors extracted from each 3D patch.
\begin{equation}
\small
{Z}_i^{{R}}=\sum_{j=1}^{N_i^R} a_{i, j}^R {F}_{i, j}^{{R}}~,~~~~
a_{i, j}^R=\frac{\exp \left\{{w}^{\top} \tanh \left({V} {F}_{i, j}^{{R}^{\top}}\right)\right\}}{\sum_{j=1}^{N_i^R} \exp \left\{{w}^{\top} \tanh \left({V} {F}_{i, j}^{{R}^{\top}}\right)\right\}}
  \label{eq23}
\end{equation}
where $a_{i, j}^R$ is the attention score predicted by the self-attention network with learnable parameters ${w}$ and ${V}$, reflecting each instance's contribution during aggregation \cite{ilse2018attention}. Its flexibility is demonstrated by the fact that the number of instances per patient input does not need to be strictly equal. Ultimately, we obtain the attention aggregated imaging features ${Z}_i^{{R}}$, pathology WSI features $W_i$, pathology report features $P_i$, and other medical report features $Z_i^B$, respectively.

\noindent\textbf{Inter-modality Aggregation}:
All modal features of a patient are aggregated into a single-dimensional feature vector. Given that a patient's modal quantity varies, we first concatenate all modal features together to form the modal feature vector set $F_i^M$. Then, we aggregate $F_i^M$ in the same manner as described in Eq.~\ref{eq23}, resulting in the final patient-level feature vector $Z_i^M \in \mathbb{R}^{1 \times 256}$.

\subsection{Patient-wise Contrastive Alignment Learning}
\label{PWCA}
We aim to create a unified joint embedding space for all modalities, facilitating more distinctive representations and the alignment of features from different modalities. It can also suppress the challenge of missing modalities when performing inter-modality aggregation. Nonetheless, we face the challenge of enforcing the alignment of multiple modalities.

This work utilizes pathology images and corresponding reports (encoded and aligned using the pre-trained PLIP model) as the hub. The strategy segregates the contrastive learning process into separate image and text sides, i.e., aligning embeddings of other visual modalities with pathology images and other textual modalities with pathology reports. The InfoNCE Loss \cite{oord2018representation} which compares the similarity of samples and encourages the model to identify positive samples among the negatives is applied for each alignment.
Due to the issue of missing modalities for each patient and the varying number of instances within each modality, it results in a batch size of one (patient) during specific training sessions. Inspired by MOCO \cite{he2020momentum}, Memory Queues have been constructed on both the image and text sides to provide a substantial number of negative samples, as shown in Fig.~\ref{fig:overview}.B.

Specifically, we first import the paired radiology image and pathology WSI of $M$ patients as the initial memory queue, denoted as $Queue_{img}$; and the paired pathological report data and other medical report data as the initial $Queue_{text}$. Samples in the memory  
queue will be popped out as the inter-patient negative samples for contrastive learning and refilled whenever new patient data are processed for training.

The process of contrastive learning on both the image and text sides is similar. Here, we take the image side as an example. 
First, we use an Adapter (two-layer Fully-connected layer \cite{gao2024clip}) to map the image features $Z_i^R$ after attention aggregation, obtaining features $R_i$. Then, the current patient's WSI features $W_i$ and $R_i$ (as a pair) are added to $Queue_{img}$, and one patient's feature pair in the queue is discarded. Next, for all feature pairs $(W,R)$ in $Queue_{img}$, with the current patient's $W_i$ and $R_i$ as the positives and other combinations as negatives, we can compute the InfoNCE Loss $L_{W, R}$.
Similarly, we use an Adapter to map the features $Z_i^B$ of other medical reports after attended aggregation, obtaining features $B_i$. Then, based on $(P,B)$, we construct $L_{P,B}$. 

The overall form of the contrastive loss is defined as $
L_{c o n}=L_{W, R}+\lambda_{con} L_{P, B}
$, where $\lambda_{con}$ is a balancing coefficient determined by the ratio of the number of complete feature pairs $(W,R)$ and $(P,B)$ among all patients.

\subsection{Progressive Survival Disambiguation Learning}
\label{PSDL}
Survival analysis is challenging because it involves ordinal regression to model time-to-event (e.g., death) data, with some events potentially not observed (right-censored). 
Following \cite{chen2021multimodal,zhou2023cross}, we divided survival times of uncensored patients into set intervals (e.g., \{0,1,2,3\}) as discrete labels for all patients, forming a maximum likelihood loss from these labels. A two-layer Prediction Layer is added to regress the death hazard and survival probability for each interval. 
For uncensored patients with accurate labels, we maximize their risk of death and minimize their survival probability in the actual discrete interval of death. 
For censored patients without accurate labels, previous works \cite{chen2021multimodal,zhou2023cross,hou2023hybrid} could only maximize survival probability in the current interval while neglecting their hazard loss.

Therefore, we propose Progressive Survival Disambiguation Learning to address this challenge by estimating unknown interval hazards for censored patients during training.
Specifically, we first use the Prediction Layer to predict risks for each interval and then use censored time to retain the hazards for subsequent intervals while setting the hazards for earlier intervals to zero. Softmax is applied to normalize these predictions, and they  
are then used to weight the ground-truth label to form the soft labels for training.
See Fig.~\ref{fig:overview}.C. for an example. 
Considering the network's limited predictive capability at the beginning of training, we further employ a time-dependent Gaussian warming up function~\cite{luo2022semi}, $\lambda_{\text {pro }}(t)=0.1 \cdot e^{\left(-5\left(1-\frac{t_i}{t_{\text {total }}}\right)^2\right)}$, to weight on these pseudo labels increasingly. Here,\ $t_i$ and $t_{\text {total }}$ denote the current and total iterations. 

The survival loss is defined as
$
L_{surv}=L_{uncen}+\lambda_{cen}\left(L_{cen}+\lambda_{pro}(t) L_{cen\_p}\right)
  $
, where $L_{uncen}$, $L_{cen}$, and $L_{cen\_p}$, which represent the loss for uncensored patients, censored patients, and the proposed risk probability estimation loss for censored patients, respectively. 
$\lambda_{cen}$ is the weight coefficient for the overall loss of censored patients. 

\section{Experiments} 
\subsection{Datasets and Tasks}
We assessed our algorithm's performance through two real-world in-house datasets, each partitioned into training and test sets on a per-patient basis for five-fold cross-validation, with mean outcomes reported. Notably, all patients have WSIs and pathology reports, while radiology data and additional clinical notes are intermittently absent. Dataset 1 contains 367 patients, of which 180 are with radiology images, 303 medical history reports, 205 colonoscopy reports, and 89 radiology reports. Dataset 2 includes 193 patients, with 133 having radiology images, 181 with medical history reports, 154 with colonoscopy reports, and 129 with radiology reports. In both, we assessed two important prognosis tasks: overall survival (OS) prediction and disease-free survival (DFS) prediction.

\subsection{Evaluation Metrics and Compared Methods}
The evaluation of our model employed three metrics. Firstly, the Concordance Index (CI) served as the primary metric, quantifying the proportion of patient pairs whose survival risks are accurately ranked. CI values span from 0 to 1, with higher values indicating superior model performance. Secondly, the Brier Score (BS) assessed prediction accuracy, calculating the mean squared difference between observed survival statuses and predicted probabilities, where a BS of 0 represents optimal accuracy. Lastly, Kaplan-Meier (KM) analysis determined patient stratification efficacy by dividing patients into high-risk and low-risk groups based on the median prediction model scores for each cohort. Superior stratification is reflected by lower p-values in the Logrank test.

Given that our data are real clinical data with multimodal missing situations in both training and testing, we primarily compared against the current SOTA multimodal algorithms for missing modalities, including MMD~\cite{cui2022survival}, HGCN~\cite{hou2023hybrid}, and ShaSpec~\cite{wang2023multi}. To ensure a fair comparison, we used the same FMs to extract features for different modalities and maintained exactly the same data division.

\begin{table}[t]
\centering
\caption{Performance comparison on the two tasks across the two datasets.}
\resizebox{0.7\linewidth}{!}{
\begin{tabular}{c|cccc|cccc}
\hline
Dataset           & \multicolumn{4}{c|}{Dataset 1}                                                         & \multicolumn{4}{c}{Dataset 2}                                                          \\ \hline
Task              & \multicolumn{2}{c|}{OS}                              & \multicolumn{2}{c|}{DFS}        & \multicolumn{2}{c|}{OS}                              & \multicolumn{2}{c}{DFS}         \\ \hline
Method            & CI $\uparrow$           & \multicolumn{1}{c|}{BS $\downarrow$}           & CI $\uparrow$           & BS $\downarrow$           & CI $\uparrow$           & \multicolumn{1}{c|}{BS $\downarrow$}           & CI $\uparrow$           & BS $\downarrow$           \\ \hline
MMD (MICCAI'22)   & 0.877          & \multicolumn{1}{c|}{0.108}          & 0.888          & 0.104          & 0.893          & \multicolumn{1}{c|}{0.102}          & 0.884          & 0.105          \\
HGCN (TMI'23)     & 0.882          & \multicolumn{1}{c|}{0.105}          & 0.891          & 0.101          & 0.899          & \multicolumn{1}{c|}{0.101}          & 0.891          & 0.102          \\
ShaSpec (CVPR'23) & 0.889          & \multicolumn{1}{c|}{0.104}          & 0.896          & 0.100          & 0.905          & \multicolumn{1}{c|}{0.098}          & 0.899          & 0.097          \\ \hline
\textbf{Ours}     & \textbf{0.897} & \multicolumn{1}{c|}{\textbf{0.100}} & \textbf{0.905} & \textbf{0.095} & \textbf{0.914} & \multicolumn{1}{c|}{\textbf{0.083}} & \textbf{0.907} & \textbf{0.091} \\ \hline
\end{tabular}}
\label{tab1}
\end{table}

\begin{table}[t]
\centering
\caption{Results of ablation studies.}
\begin{minipage}{0.3\linewidth}
\centering
\resizebox{0.8\linewidth}{!}{
\begin{tabular}{c|cc}
\hline
Method                & CI $\uparrow$           & BS $\downarrow$           \\ \hline
Baseline          & 0.871          & 0.109          \\
w/o $L_{con}$  & 0.884          & 0.105          \\
w/o $L_{cen\_p}$        & 0.887          & 0.104          \\ \hline
\textbf{Ours}         & \textbf{0.897} & \textbf{0.100} \\ \hline
\end{tabular}}
\end{minipage}%
\hfill
\begin{minipage}{0.3\linewidth}
\centering
\resizebox{0.75\linewidth}{!}{
\begin{tabular}{c|cc}
\hline
Method            & CI $\uparrow$           & BS $\downarrow$           \\ \hline
WSI   & 0.856          & 1.116          \\
$P_{rep}$ & 0.842          & 1.119          \\
WSI$\&$$P_{rep}$  & 0.865          & 1.111          \\ \hline
\textbf{Ours}     & \textbf{0.897} & \textbf{0.100} \\ \hline
\end{tabular}}
\end{minipage}%
\hfill
\begin{minipage}{0.3\linewidth}
\centering
\resizebox{0.7\linewidth}{!}{
\begin{tabular}{c|cc}
\hline
Method        & CI $\uparrow$           & BS $\downarrow$           \\ \hline
PLIP     & 0.889          & 0.104          \\
CLIP     & 0.885          & 0.105          \\ \hline
\textbf{Ours} & \textbf{0.897} & \textbf{0.100} \\ \hline
\end{tabular}}
\end{minipage}%
\label{tab2}
\end{table}

\begin{figure}[t!]
  \centering
   \includegraphics[width=\linewidth]{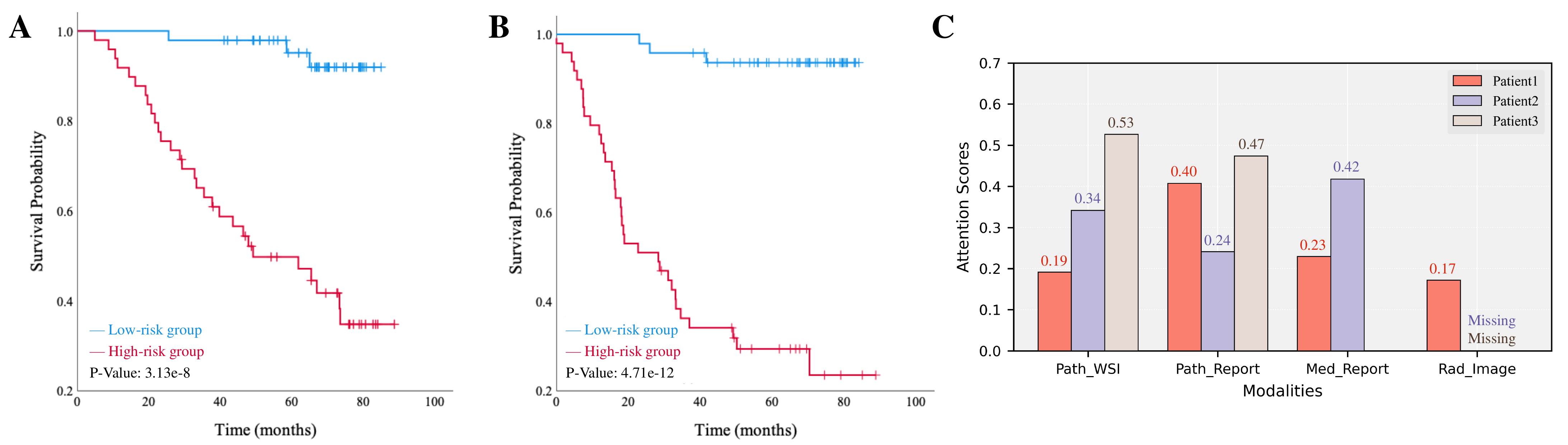}
   \caption{The KM analysis curves for (A) OS prediction task and (B) DFS prediction task. (C) the visualization of modal attention scores for three patients.}
   \label{fig:high-low}
\end{figure}

\subsection{Results}
As shown in Table~\ref{tab1}, the performance of our algorithm significantly surpasses all comparison algorithms across all tasks in all datasets, indicating the high efficiency of our algorithm.

We further divided the test cohort of Dataset 1 into high-risk and low-risk groups based on the median risk score predicted by our model. If our model's predictions are efficient, then there should be a significant difference between the KM curves of these two groups. The results, as shown in Fig.~\ref{fig:high-low}.A (OS prediction task) and Fig.~\ref{fig:high-low}.B (DFS prediction task), reveal that the p-values for both groups are less than 1e-7, indicating the significant efficacy of our method.

Our method also possesses strong clinical interpretability, allowing for the flexible quantification of the importance of each modality engaged by the test patients toward the outcomes. Fig.~\ref{fig:high-low}.C illustrates the visualization of modal attention scores for three typical patients, demonstrating our method's robust and flexible interpretative advantage in clinical applications.

\noindent\textbf{Ablation Study}:
In our detailed ablation studies for the Overall Survival (OS) prediction task using Dataset 1, we explored three different aspects, with results presented in Table~\ref{tab2}. First, we conducted ablation experiments on the proposed contrastive learning loss $L_{con}$, and the loss $L_{cen\_p}$ tailored for censored data. The baseline represents scenarios without any contrastive learning or disambiguation learning, highlighting the effectiveness of these two key components we introduced. Second, we performed ablation experiments with pathology WSIs ($WSI$) only,  pathology reports ($P_{rep}$) only, and both, demonstrating that the joint learning of multiple modalities can effectively enhance performance. Third, we compared the performance of using a single vision-language large model, either PLIP~\cite{huang2023visual} or CLIP~\cite{radford2021learning}, for all tasks. The results indicated that using a specialized medical large model for each modality can improve performance.

\section{Conclusion}
Our paper presents a novel multi-modal survival analysis framework tailored to address critical challenges in cancer treatment research, including incomplete data and censored survival labels. It represents a significant advancement in leveraging multi-modal data and overcoming critical challenges for precise survival predictions in cancer treatment research. Future research avenues include exploring application and validation in large-scale multi-center studies.

\section*{Acknowledgement}
This work is funded by the National Key R\&D Program of China (2022ZD0160700) and Shanghai AI Laboratory.

\section*{Disclosure of Interests}
We declare no competing interests.

%
%
%
\bibliographystyle{splncs04}
\bibliography{reference}
\end{document}